\pdfoutput=1

\documentclass[11pt]{article}

\usepackage[]{eacl2023}
\usepackage{amsfonts,amssymb}
\usepackage{hyperref}
\usepackage{times}
\usepackage{latexsym}
\usepackage{graphicx} 
\usepackage{booktabs}

\usepackage[T1]{fontenc}
\usepackage{amsmath}
\usepackage{bbding}
\usepackage{pifont}
\newcommand{\cmark}{\ding{51}}%
\newcommand{\xmark}{\ding{55}}%
\usepackage{caption}
\usepackage{subcaption}
\usepackage{comment}
\usepackage{color}

\usepackage[utf8]{inputenc}

\usepackage{microtype}
\usepackage{cleveref}

\title{Parameter-Efficient Tuning with Special Token Adaptation}

\usepackage{xspace}
\usepackage{cleveref}
\crefformat{section}{\S#2#1#3}
\crefformat{subsection}{\S#2#1#3}
\crefformat{subsubsection}{\S#2#1#3}
\crefrangeformat{section}{\S\S#3#1#4 to~#5#2#6}
\crefmultiformat{section}{\S\S#2#1#3}{ and~#2#1#3}{, #2#1#3}{ and~#2#1#3}
\usepackage{refstyle}
\Crefformat{figure}{#2Fig.~#1#3}
\Crefmultiformat{figure}{Figs.~#2#1#3}{ and~#2#1#3}{, #2#1#3}{ and~#2#1#3}
\Crefformat{table}{#2Tab.~#1#3}
\Crefmultiformat{table}{Tabs.~#2#1#3}{ and~#2#1#3}{, #2#1#3}{ and~#2#1#3}
\Crefformat{appendix}{Appx.~\S#2#1#3}
\crefformat{algorithm}{Alg.~#2#1#3}

\newcommand{\modelname}{\textsc{PaSTA}\xspace}
\newcommand{\cls}{\texttt{[CLS]}\xspace}
\newcommand{\sep}{\texttt{[SEP]}\xspace}

\newcommand{\stitle}[1]{\vspace{0.3ex} \noindent{\bf #1}}

\author{
Xiaocong Yang$^{\dagger}$\Thanks{Work done when visiting
USC.}, James Y. Huang$^{\ddagger}$, Wenxuan Zhou$^{\ddagger}$ \and Muhao Chen$^{\ddagger}$\\
$^{\dagger}$Tsinghua University;\;
$^{\ddagger}$University of Southern California\\ 
\texttt{yangxc.18@sem.tsinghua.edu.cn};\;\texttt{\{huangjam,zhouwenx,muhaoche\}@usc.edu}\\
}

\begin{document}
\maketitle
\begin{abstract}
Parameter-efficient tuning aims at updating only a small subset of parameters when adapting a pretrained model to downstream tasks. In this work, we introduce \modelname, in which we only modify the special token representations (e.g., \texttt{[SEP]} and \cls in BERT) before the self-attention module at each layer in Transformer-based models. \modelname achieves comparable performance to full finetuning in natural language understanding tasks including text classification and NER with up to only 0.029\% of total parameters trained. Our work not only provides a simple yet effective way of parameter-efficient tuning, which has a wide range of practical applications when deploying finetuned models for multiple tasks, but also demonstrates the pivotal role of special tokens in pretrained language models.\footnote{Our code is publicly available at:  \texttt{\href{https://github.com/luka-group/PASTA/}{\url{https://github.com/luka-group/PASTA/}}} }

\end{abstract}

\section{Introduction}

Built upon a pretrained language model (PLM; \citealt{BERT,Roberta,XLNet,Palm}), many of the recent NLP systems are developed based on task-specific finetuning. 
In this way, the PLM effectively leverages the task-agnostic knowledge captured during self-supervised pretraining and adapts itself to downstream tasks. 
However, full finetuning poses a challenge to model deployment under multi-task, memory-limited scenarios, where we need to train and store a separate full-sized model for each substantially distinct task. 
As an alternative, parameter-efficient tuning~\cite{Delta-Tuning} aims at only updating a small number of parameters when adapting PLMs to downstream tasks while making most of the model parameters fixed and shared among tasks, thus reducing memory usage. 

\begin{figure}

\centering
\begin{subfigure}{0.1\textwidth}
\centering
\includegraphics[width=\textwidth]{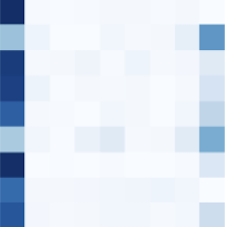}
\caption*{L5-H3}
\end{subfigure}
\begin{subfigure}{0.1\textwidth}
\centering
\includegraphics[width=\textwidth]{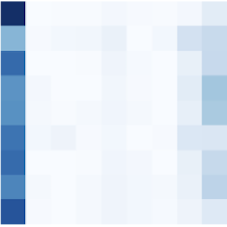}
\caption*{L5-H5}
\end{subfigure}
\begin{subfigure}{0.1\textwidth}
\centering
\includegraphics[width=\textwidth]{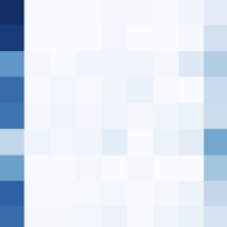}
\caption*{L5-H6}
\end{subfigure}
\begin{subfigure}{0.1\textwidth}
\centering
\includegraphics[width=\textwidth]{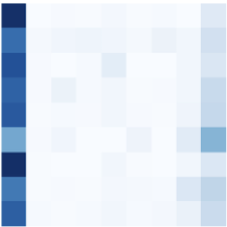}
\caption*{L5-H7}
\end{subfigure}

\centering
\begin{subfigure}{0.1\textwidth}
\centering
\includegraphics[width=\textwidth]{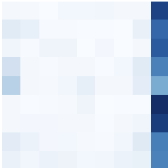}
\caption*{L20-H3}
\end{subfigure}
\begin{subfigure}{0.1\textwidth}
\centering
\includegraphics[width=\textwidth]{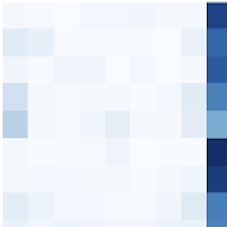}
\caption*{L20-H5}
\end{subfigure}
\begin{subfigure}{0.1\textwidth}
\centering
\includegraphics[width=\textwidth]{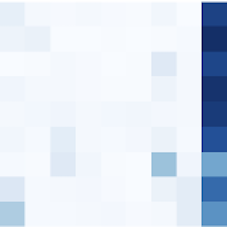}
\caption*{L20-H8}
\end{subfigure}
\begin{subfigure}{0.1\textwidth}
\centering
\includegraphics[width=\textwidth]{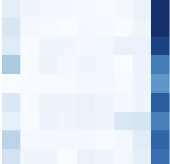}
\caption*{L20-H15}
\end{subfigure}
\caption{Examples of vertical attention heads in the 5-th and 20-th layer of BERT-large with a random sample from CoLA \cite{warstadt-etal-2019-neural} as input. Heads in the first row and second row assign most of maximal attention weights to \texttt{[CLS]} and \texttt{[SEP]} respectively. See \Cref{sec:full_map} for the full attention map.}
\label{fig:main_attention}
\end{figure}

In this paper, we propose \textbf{\textsc{Pa}}rameter-efficient tuning with \textbf{\textsc{S}}pecial \textbf{\textsc{T}}oken \textbf{\textsc{A}}daptation (\modelname),  where we only add trainable vectors to hidden representations of \emph{special tokens} \footnote{WLOG, we use the notation of special tokens \cls and \sep in BERT for the convenience of expression,
while the method applies in the same way to other paradigms such as \texttt{<S>} and \texttt{</S>} in RoBERTa \cite{Roberta}.} at each layer before the multi-head attention module in Transformer-based PLMs.
Our work is motivated by the role of special tokens in PLMs.
First, special tokens such as \cls collect information from the whole input sequence and are typically regarded as the global text representation~\cite{BERT}. 
For sentence-level tasks such as GLUE~\cite{GLUE}, a common practice is to add a new classifier head based on the \cls representation in the last model layer.
Thus, if trained properly, by updating the \cls representations, we can approximate the result of the \textit{information collection} process in PLMs.
Second, many attention heads in PLMs follow a vertical pattern\footnote{\label{fn:vertical}Following \citet{voita-etal-2019-analyzing} and \citet{TLM}, an attention head is regarded as \textit{vertical} if at least 90\% tokens assign maximal attention scores to either \cls or \sep.}, where the attention scores are mostly allocated to either the \cls or \sep token~\cite{lookat, reveal}, as illustrated in \Cref{fig:main_attention}.
Therefore, updates to special tokens can also be \textit{disseminated} to other tokens during the forward pass through the vertical attention heads~\cite{Mathmatical_framework}, enabling the PLMs to adapt to both sentential and lexical tasks.

By tuning as few as up to 0.029\% of the total parameters, \modelname achieves competitive performance on par with full finetuning and BitFit \cite{BitFit} on GLUE (\Cref{ssec:glue}). It also outperforms P-tuning v2 \cite{p-tuningv2} by 0.6\% on CoNLL2003 with 20$\times$ fewer additional parameters (\Cref{ssec:ner}). 
The ablation study shows that we can further reduce trainable parameters to 0.009\% with only a slight performance drop (\Cref{sec:analysis}), 
showing the merit of adapting special token representations.


\section{Related Work}


A recent survey~\cite{Delta-Tuning} categorizes 
three types of \emph{parameter-efficient tuning} methods. 
\emph{Addition methods}~\cite{Adapter,prompt-tuning,p-tuningv2} introduce a small number of additional trainable parameters while keeping those in the PLM unchanged. 
\emph{Specification methods}~\cite{BitFit,Diff-pruning, Masking} update a portion of parameters in the PLM while keeping others frozen. 
\emph{Reparameterization methods}~\cite{intrinsic,LoRA,Uni_intrinsic} modify PLMs' structures to parameter-efficient forms.
Our method belongs to the addition-based methods and follows the basic settings of 
P-tuning~v2~\cite{p-tuningv2}, where newly initialized hidden representations of tokens are inserted into each Transformer layer. Different from most prompt tuning methods that introduce new tokens, we add the introduced vectors to the hidden states of special tokens and keep the sequence length unchanged.


Previous works use probing tasks~\cite{ parameterfree} and pruning methods~\cite{Lottery} to study the roles of different modules inside BERT.
It has been shown that functional specialization exists in BERT self-attention heads~\cite{lookat}, and vertical attention heads\textsuperscript{\ref{fn:vertical}} take up a large portion~\cite{TLM}. \citet{reveal} find that vertical attention heads are almost exclusively associated with attention to \sep or \cls tokens, and \citet{lookat} conclude that heads in early layers often attend to \cls while in middle layers attend to \sep. 
In this work, we demonstrate that adapting hidden representations of special tokens is sufficient to bring the performance of PLMs to the level of full finetuning.



\section{PASTA}
\begin{figure}[t]
    \centering
    \includegraphics[width=0.47\textwidth]{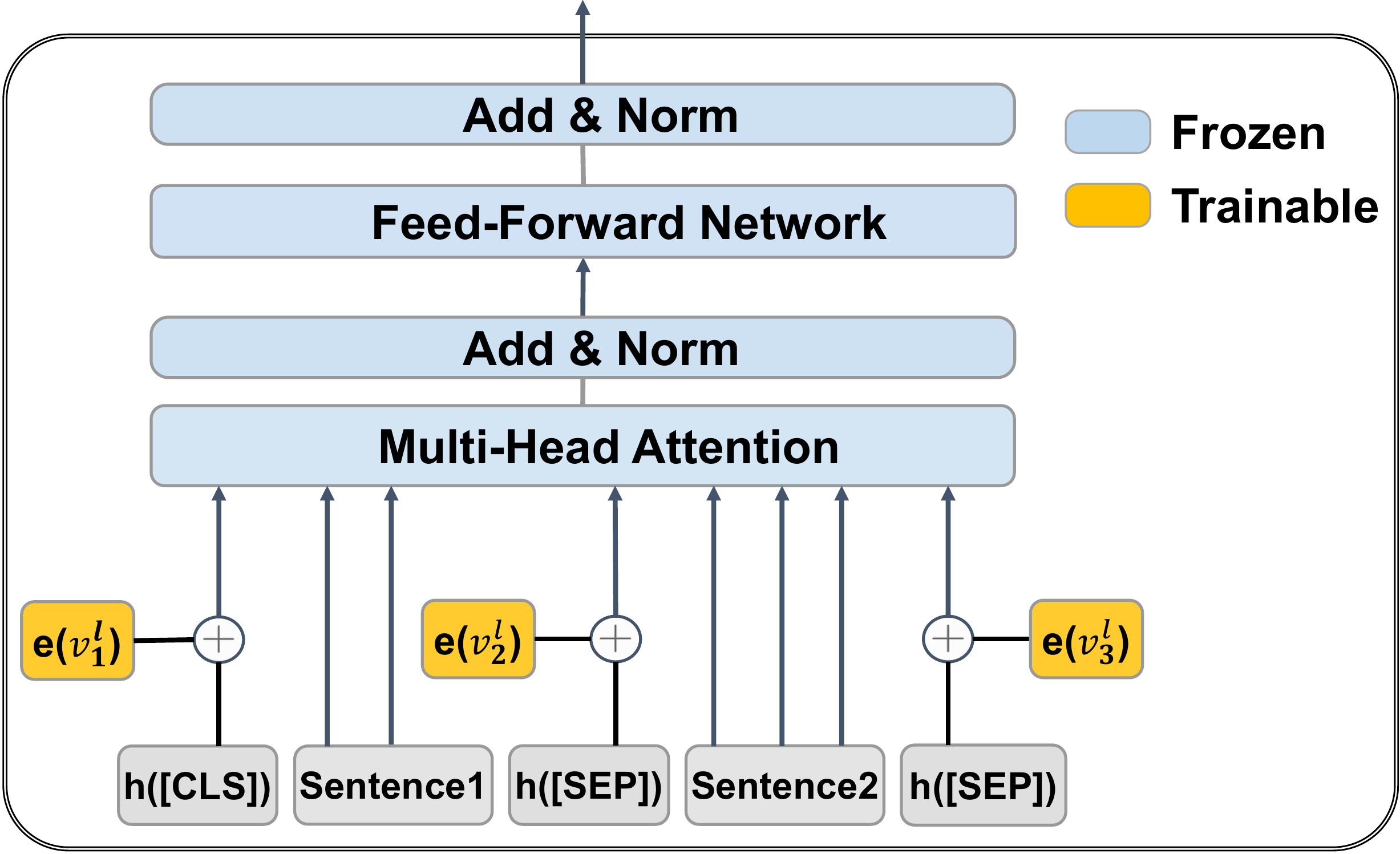}
    \caption{Architecture of \modelname layer in Transformer. Skip-connections in Transformers are not shown for brevity. At layer $l$ we add a trainable vector $\mathbf{e}(\mathbf{v}_p^l)\in \mathbb{R}^{d}$ to the hidden representation of the $p$-th special token in the input sequence, and freeze the weights of the PLM.}
    \label{fig:architect}
\end{figure}

Given a large PLM, our goal is to develop a parameter-efficient tuning method that only updates a small set of 
parameters when adapting to a downstream task. To this end, we propose a simple yet effective method called \modelname, in which we train a hidden vector for every special token at each Transformer layer, along with a task-specific classifier, while freezing the parameters of the PLM. 



\subsection{Special Token Adaptation}



The special token adaption is illustrated in~\Cref{fig:architect}.
Although these adaptations are not directly applied to non-special tokens, changes in special token hidden states can be effectively disseminated to other tokens via self-attention during forward passes, thanks to the prevalence of vertical attention heads\textsuperscript{\ref{fn:vertical}} in PLMs.

Specifically, denote the inputs to the $l$-{th} Transformer layer as $\mathbf{H}^l = \{\mathbf{h}_i^l\}_{i=1}^N, \mathbf{h}_i^l\in \mathbb{R}^d$, where $N$ is the number of input tokens, $d$ is the hidden size, \modelname modifies the inputs as follows:

\begin{align*}
    \mathbf{H}^l_\text{mod} &= \{\mathbf{h}_i^l + \mathbf{m}_i^l\}_{i=1}^N, \\
    \mathbf{H}^{l+1} &= \text{Trm}^l(\mathbf{H}^l_\text{mod}),
\end{align*}

where $\text{Trm}^l$ is the $l$-th Transformer layer, $\mathbf{m}_i^l \in \mathbb{R}^{d}$ is our special token adaptation defined as follows:

\begin{equation*}  
\mathbf{m}^l_i = \left\{ 
\begin{aligned}
\mathbf{0}  \qquad \text{if token $i$ is not a special token} \\
\mathbf{e}(\mathbf{v}_p^l) \quad \text{if token $i$ is the $p$-th special token} 
\end{aligned}
\right.
\end{equation*}

\noindent
where $\mathbf{e}(\mathbf{v}_p^l) \in \mathbb{R}^d$ is the trainable vector added to the hidden representation of the $p$-th special token in the input sequence. 
During 
downstream task training, only those introduced hidden vectors for special tokens and the task-specific classifier are 
optimized, and the rest of model parameters are frozen.






\subsection{Parameter Efficiency and Consistency}
As shown in \Cref{tab:property},
\modelname achieves $\mathcal{O}(L\times d)$ parameter complexity\footnote{The number of special tokens are invariant to the scale of models, and are usually very small.} and updates as few as 0.015\%-0.029\% of the parameters compared to a full PLM when using BERT-large or RoBERTa-large as backbone. Unlike Adapter~\cite{Adapter} that learns the transformation of all input tokens using a shared FFN, \modelname only learns the task-specific update of special token representations as a bias term, 
which significantly reduces the parameter capacity needed for adaptation on downstream tasks. 

Meanwhile, the set of parameters introduced by \modelname is consistent across different tasks, making it efficient for hardware-based deployment~\cite{BitFit}. On the contrary, in Diff-Prune, the parameter update is considered as a term of the loss function~\cite{Diff-pruning}, resulting in different sets of updated parameters in distinct tasks.

\begin{table}[t]
\small
\centering
\setlength{\tabcolsep}{5pt}
\begin{tabular}{lcc}
\toprule
\textbf{} & \textbf{\# Param} & \textbf{Parameter Consistency} \\
\midrule
Adapter  & $\mathcal{O}(L\times d \times r)$ & \cmark \\
P-tuning v2  & $\mathcal{O}(L\times d \times T)$ & \cmark  \\
BitFit  & $\mathcal{O}(L\times (d+m))$ & \cmark  \\
Diff-Prune & -  & \xmark  \\
\midrule
\modelname & $\mathcal{O}(L\times d)$ & \cmark  \\
\bottomrule
\end{tabular}
\caption{Parameter complexity of \modelname and baselines. Here $L$ and $d$ refer to the number of layers and hidden size of the PLM.  $m$ and $r$ refer to the intermediate size of FFN modules in Transformers and Adapters, respectively. $T$ is the prompt length. Parameter consistency refers to whether the set of trainable parameters is consistent across different tasks \cite{BitFit}.}
\label{tab:property}
\end{table}

\section{Experiments and Results}\label{sec:exp}

We hereby study the downstream performance of \modelname and analyze the properties of introduced hidden vectors.

\begin{table*}[t]
\small
\centering
\setlength{\tabcolsep}{4.2pt}
\begin{tabular}{lccccccccccc}
\toprule
\textbf{} & \textbf{\%Param} & \textbf{RTE} & \textbf{CoLA} & \textbf{STS-B}& \textbf{MRPC}& \textbf{SST-2}& \textbf{QNLI} &\textbf{MNLI(m/mm)} &\textbf{QQP} & \textbf{Avg.}\\
\textbf{} & &acc. & mcc. & Spearman& F1 &acc.&acc.&acc.&F1&\\
\midrule
Full Finetuning$^{\ast}$ & 100\% & 70.1 & 60.5  &86.5  & 89.3 & \textbf{94.9} & 92.7 & \textbf{86.7/85.9} & \textbf{72.1} & \textbf{81.6}\\
Adapter$^{\ast\ast}$  & 3.6\% & 71.5 &59.5  &\textbf{86.9}  & 89.5 & 94.0 & 90.7 & 84.9/85.1 &  71.8 & 81.1\\
Diff-Prune$^{\dag}$  & 0.5\%& 70.6 & 61.1 &86.0  &\textbf{89.7}  & 94.1 & \textbf{93.3} & 86.4/86.0 &  71.1 & 81.5\\
P-tuning v2 & 0.29\% & 70.1 & 60.1 & 86.8 & 88.0
 & 94.6 & 92.3 & 85.3/84.9 & 70.6 & 81.0 \\
BitFit$^{\ddag}$  & 0.08\% & \textbf{72.0} &59.7  &85.5  & 88.9 & 94.2 & 92.0 & 84.5/84.8 &  70.5 & 80.9 \\

\midrule
\modelname & \textbf{0.015\%-0.022\%} & 70.8 & \textbf{62.3} & 86.6 & 87.9 & 94.4 &  92.8 & 83.4/83.4   & 68.6 & 80.9  \\
\bottomrule
\end{tabular}
\caption{BERT-large model performance on GLUE benchmark test set. Lines with $^\ast$ and $^{\ast\ast}$ are results from \citet{BERT} and \citet{Adapter}, and lines with $^\dag$ and $^\ddag$ are from \citet{Diff-pruning} and \citet{BitFit} respectively. We reimplement experiments of P-tuning v2 on GLUE benchmark with a prompt length of 20.}
\label{tab:glue}
\end{table*}
\begin{table*}[t]
\small
\centering
\setlength{\tabcolsep}{4.2pt}
\begin{tabular}{lccccccccccc}
\toprule
\textbf{} & \textbf{\%Param} & \textbf{RTE} & \textbf{CoLA} & \textbf{STS-B}& \textbf{MRPC}& \textbf{SST-2}& \textbf{QNLI} &\textbf{MNLI(overall)} &\textbf{QQP} & \textbf{Avg.}\\
\textbf{} & &acc. & mcc. & Pearson & acc. &acc.&acc.&acc.&acc.&\\
\midrule
Full Finetuning$^{\ast}$ & 100\% & 86.6 & 68.0 & 92.4 & \textbf{90.9} & 96.4 & 94.7 & 90.2 & \textbf{92.2} & 88.9\\
LoRA$^{\dag}$ & 0.24\% & \textbf{87.4} & 68.2 & \textbf{92.6} & \textbf{90.9} & 96.2 & 94.9 & \textbf{90.6} & 91.6 & \textbf{89.0} \\ 

\midrule
\modelname & \textbf{0.015\%-0.029\%} & 86.6 & \textbf{69.7} & 91.8 & \textbf{90.9} & \textbf{96.8} & \textbf{95.1} & 90.4  & 89.9 & 88.9  \\
\bottomrule
\end{tabular}
\caption{RoBERTa-large model performance on GLUE benchmark. Lines with $^\ast$ are results from \citet{Roberta}, and lines with $^\dag$ are from \citet{LoRA}. We follow the metric settings of baselines and also report results on GLUE development set for the convenience of direct comparison.}
\label{tab:roberta}
\end{table*}

\begin{table}[t]
\small
\centering
\setlength{\tabcolsep}{2pt}
\begin{tabular}{lccccc}
\toprule
\textbf{} & \textbf{CoLA} & \textbf{RTE} & \textbf{MRPC} & \textbf{STS-B}& \textbf{CoNLL2003}\\
\midrule
\modelname  & \textbf{65.4}& \textbf{76.2} & 89.7 & \textbf{90.8} & \textbf{94.0} \\
- w/o \texttt{[CLS]} & 58.8 & 72.6 & 91.4 & 90.2 &93.7 \\
- w/o \texttt{[SEP]} & 64.5 & 71.1 & 91.9 & 90.3  & 93.7\\
- shared vector & 64.7 & 74.7 & \textbf{92.1} & 90.0 & 93.9 \\
- classifier only & 36.5 & 54.2 & 81.5 & 64.9 & 77.4 \\
\bottomrule
\end{tabular}
\caption{Performance of ablation study with BERT-large on GLUE and CoNLL2003 development sets. }
\label{tab:ablation}
\end{table}

\subsection{Experimental Setup}

\stitle{Baseline Methods.}
We compare \modelname with the following parameter-efficient tuning methods in prior studies. 
\textbf{Adapter}~\cite{Adapter} introduces new feed-forward modules in Transformer layers while keeping original parameters of the PLM frozen. \textbf{BitFit}~\cite{BitFit}  updates all bias terms in the PLM during finetuning. \textbf{Diff-Prune} \cite{Diff-pruning} introduces $L_0$-norm penalty on the updated parameters to encourage sparsity of tuned parameters. \textbf{P-tuning v2} \cite{p-tuningv2} prepends trainable hidden vectors before the input sequence at each layer while keeping the original PLM parameters frozen. \textbf{LoRA} \cite{LoRA} uses low-rank decomposition matrices to model the parameter updates.

\stitle{Model Configuration.} We conduct our experiments using BERT-large and RoBERTa-large (We also report experiments with BERT-base in \Cref{sup:base}). To facilitate comparison with baseline works, we take most of the experimental results from their original papers which are reported with either BERT-large or RoBERTa-large. Note that multiple \sep  tokens in a single sequence (e.g., in sentence pair tasks like MNLI) are treated as different special tokens and have separate sets of trainable parameters, and the number of trainable parameters varies among downstream tasks according to the number of special tokens added. Details of training and hyperparameters settings are shown in \Cref{sec:details}.

\subsection{GLUE Tasks}\label{ssec:glue}
\stitle{Task Setup.}
We evaluate \modelname on the widely used GLUE benchmark\footnote{Following previous work \cite{Adapter,Diff-pruning,BitFit}, we exclude WNLI since BERT underperforms the majority class baseline \cite{BERT}.}~\cite{GLUE}. 
For the convenience of direct comparison, we use the same metrics as were used in baseline works \cite{BERT, Roberta}. For experiments with BERT, MRPC and QQP are evaluated using F1 score, STS-B is evaluated using Spearman’s correlation coefficient, CoLA is evaluated using Matthew’s Correlation, and the other tasks are evaluated using accuracy. For experiments with RoBERTa, STS-B is evaluated using Pearson’s correlation coefficient, CoLA is evaluated using Matthew’s Correlation, and the other tasks are evaluated using accuracy.

\stitle{Results.} 
\Cref{tab:glue,tab:roberta} report the performance of \modelname on GLUE benchmark with BERT-large and RoBERTa-large respectively. 
\modelname with RoBERTa-large achieves the same average score as that of full finetuning over GLUE tasks. \modelname with BERT-large achieves an average score on par with BitFit using over 3$\times$ fewer trainable parameters and comparable results to other higher parameter complexity baselines. 
The results demonstrate that by leveraging the pivotal role of special tokens in PLMs,
\modelname is able to effectively adapt the model to sentence-level tasks with significantly fewer parameters tuned than previous methods.

\subsection{Named Entity Recognition}\label{ssec:ner}
\stitle{Task Setup.}
We experiment with the NER task on CoNLL2003~\cite{tjong-kim-sang-de-meulder-2003-introduction}.
Following~\citet{BERT}, we formulate NER as a token classification problem.

\stitle{Results.}
\begin{figure}[t]
    \centering
    \includegraphics[width=0.4\textwidth]{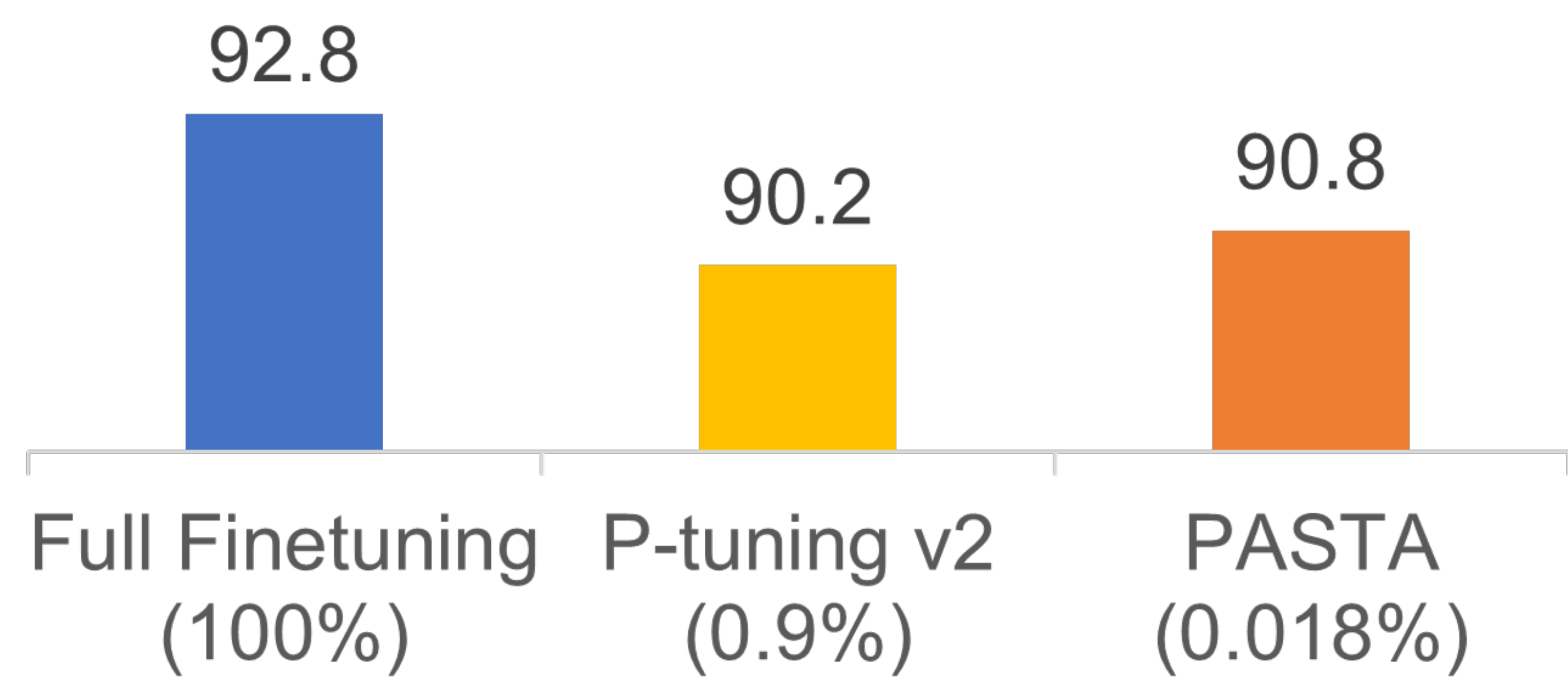}
    \vspace{-0.5em}
    \caption{NER results on CoNLL03 with BERT-large (F1 score percentages are marked over the bars). Each method is labeled with the percentage of trainable parameter sizes with regard to full finetuning in parentheses.}
    \label{fig:conll03}
    \vspace{-0.5em}
\end{figure}
As shown in \Cref{fig:conll03}, \modelname with BERT-large achieves an F1 score of 90.8\% on CoNLL2003 test set, outperforming P-tuning v2 \cite{p-tuningv2} by 0.6\% with 20$\times$ fewer trainable parameters, while falling behind full finetuning by 2.0\%. 
Nevertheless, the strong performance of \modelname compared to P-tuning v2 indicates that even though \modelname only directly adapts special tokens, the representations of all input tokens can still be properly tuned, supporting our hypothesis that vertical attention heads are able to disseminate adaptations in special token hidden states to other tokens. 

\subsection{Analysis}\label{sec:analysis}

\stitle{Ablation on choices of special tokens.}
To understand the effect of tuning different combinations of special tokens on downstream tasks, we further limit the additional parameter capacity of \modelname by only adapting either \cls or \sep, or share a common vector across all special tokens. \Cref{tab:ablation} shows the performance of three ablated variants and a baseline that only tunes the classification head on top of a fixed BERT-large. In general, we observe a decrease in performance on most tasks except on MRPC for three \modelname variants, and performance degrades significantly if we do not adapt any special tokens. These results demonstrate the vital role of introduced hidden vectors for special tokens in \modelname, while the best choice of special tokens to be adapted may vary depending on the task.

\begin{figure}[]
    \centering
    \includegraphics[width=0.47\textwidth]{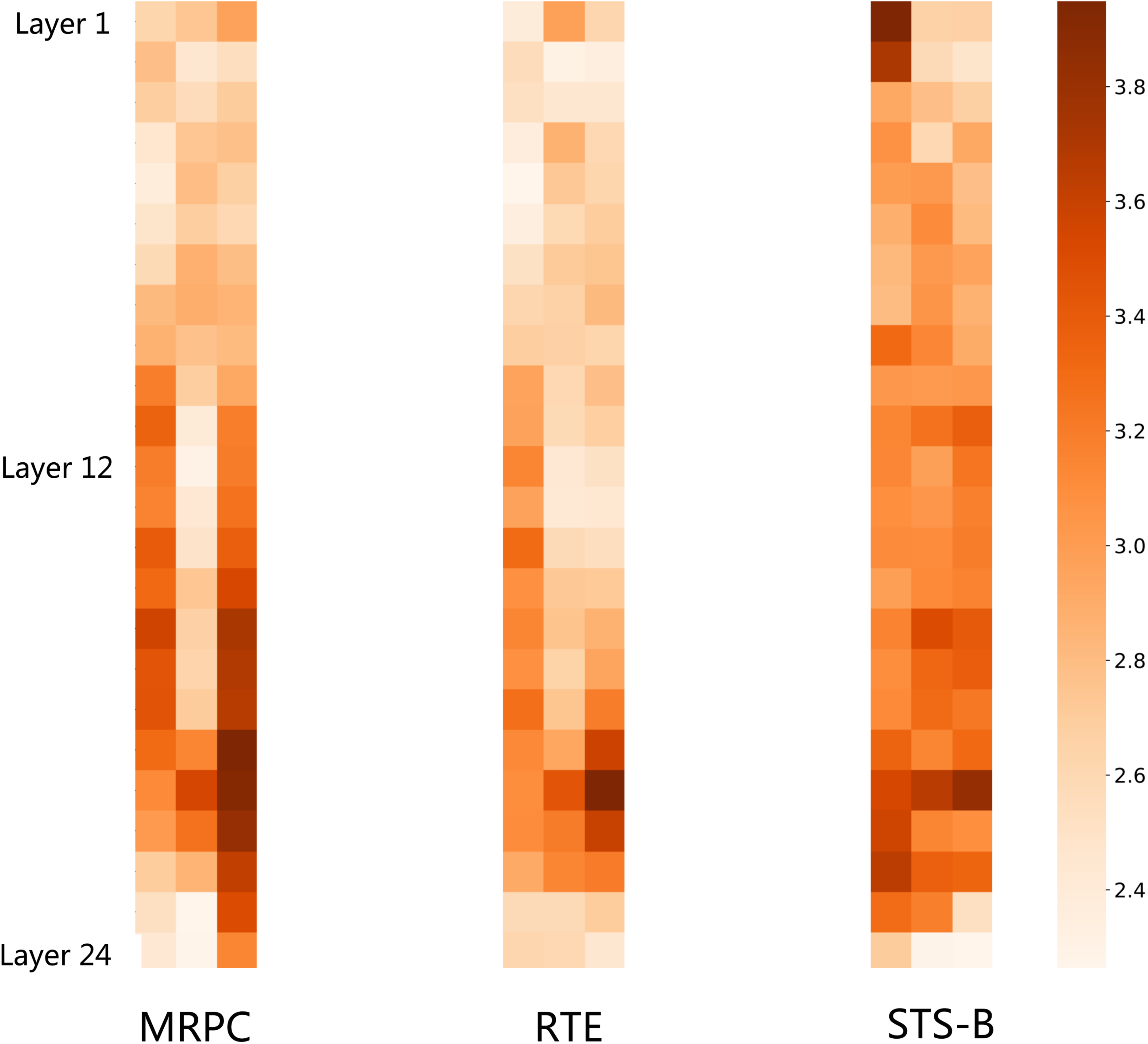}
    \caption{The distribution map of norms of introduced hidden vectors on MRPC, RTE and STS-B tasks with BERT-large. In each subgraph, the first column shows the norms of introduced vectors added to \texttt{[CLS]} at each layer, and the second and third columns are introduced vector norms at two \texttt{[SEP]} tokens respectively. }
    \label{fig:Norm}
\end{figure}

\stitle{Norm distribution of introduced hidden vectors.} \Cref{fig:Norm} shows the norm distribution of introduced vectors on downstream tasks.
The introduced hidden vectors learn the difference of special tokens between pretrained and adapted models, and thus norms of those vectors indicate the magnitude of parameter change at different layers. Similar to the pattern of parameter change during full finetuning \cite{reveal}, \modelname generally has larger norms of hidden vectors at layers closer to the output.

\section{Conclusion}

We present \modelname, a parameter-efficient tuning method that only modifies special token representations at each Transformer layer when adapting to downstream tasks. Our approach is motivated by the observation that PLMs have a large amount of vertical attention heads that heavily attend to special tokens, and these heads disseminate value updates from special tokens to all of the other tokens. Experiments show that \modelname achieves strong performance comparable to full finetuning on sentential and lexical tasks with high parameter efficiency. Our work not only provides an effective solution for parameter-efficient tuning, but also demonstrates the pivotal role of special tokens in PLMs.
\section*{Limitations}
In this work we hypothesize that the vertical attention heads could play a role as ``information disseminator'' based on the theoretical analysis of Transformers \cite{Mathmatical_framework}. However, we still have no direct approaches such as probing tasks and reverse engineering to prove this assumption. And since \modelname relies on adaptation of special tokens, it cannot be applied to language models which do not pad special tokens to input sequences such as GPT-2 \cite{Radford2019LanguageMA}.
For the empirical results, we choose GLUE benchmark and CoNLL2003 to study the performance on language understanding tasks. The effectiveness of \modelname on language generation tasks has not been tested in this work due to limited bandwidth. Finally, similar to other parameter-efficient tuning methods, \modelname suffers from a higher computational cost compared to full finetuning.

\section*{Acknowledgement}
We appreciate the anonymous reviewers for their insightful comments and suggestions. Xiaocong Yang, Wenxuan Zhou and Muhao Chen
are supported by the National Science Foundation
of United States Grant IIS 2105329. 
This material is supported in part by the DARPA MCS program under Contract No. N660011924033 with the United States Office Of Naval Research, an Amazon Research Award, a Cisco Research Award and a subaward from NSF Cloudbank 1925001.



\bibliography{reference}
\bibliographystyle{acl_natbib}

\appendix

\section{Performance of BERT-base}\label{sup:base}
\Cref{tab:base} reports the performance of \modelname and a part of baselines with BERT-base as backbones on GLUE development sets. We do not put the result table in the main body of this work because most baselines did not report GLUE test set scores using BERT-base, and only three of them reported GLUE development set scores with BERT-base in their works.
\modelname slightly underperforms baselines on average, while it outperforms other models on small datasets such as RTE and MRPC.
\begin{table*}[b!]
\small
\centering
\setlength{\tabcolsep}{5pt}
\begin{tabular}{lccccccccccc}
\toprule
\textbf{} & \textbf{\%Param} & \textbf{RTE} & \textbf{CoLA} & \textbf{STS-B}& \textbf{MRPC}& \textbf{SST-2}& \textbf{QNLI} &\textbf{MNLI(m/mm)} &\textbf{QQP} & \textbf{Avg.}\\
\midrule
Full Finetuning $^\ast$ & 100\% & 66.4 & \textbf{62.1} & \textbf{89.8} & 90.9 & 91.6 &  90.0& \textbf{83.2/ -} & \textbf{87.4} & 82.7 \\
Adapter$^\ast$ & 0.81\% & 71.8 & 61.5 & 88.6 & 89.9 & 91.9 & \textbf{90.6}& 83.1/ - & 86.8 & \textbf{83.0} \\
BitFit$\dag$ & 0.8\% & 72.3 & 58.8 & 89.2 & 90.4 & \textbf{92.1} &  90.2 & 81.4/ - & 84.0 & 82.3 \\
\midrule
\modelname & \textbf{0.015\%-0.022\%} & \textbf{73.6} &57.9 & 88.7 & \textbf{91.5} & 91.2 & 89.7 &77.8/78.8 & 80.8 & 81.4 \\
\bottomrule
\end{tabular}
\caption{PASTA with BERT-base model performance on GLUE benchmark development set. Lines with $^\ast$ and $^\dag$ refer to results from \citet{Unipelt} and \citet{BitFit} respectively.}
\label{tab:base}
\end{table*}

\section{Implementation Details}
\label{sec:details}
\begin{table*}[b!]
\small
\centering
\setlength{\tabcolsep}{6pt}
\begin{tabular}{lcccccccccc}
\toprule
\textbf{}  & \textbf{RTE} & \textbf{CoLA} & \textbf{STS-B}& \textbf{MRPC}& \textbf{SST-2}& \textbf{QNLI} &\textbf{MNLI} &\textbf{QQP} & \textbf{CoNLL2003}\\
\midrule
Learning rate  & 4.5e-3 & 5e-3 &  2e-3 & 2.5e-3 & 7e-3 & 2e-3 & 5e-4 & 5e-3 & 3e-3\\

Batch size  & 32$\times$4 & 32$\times$1 & 32$\times$3 & 32$\times$4 & 64$\times$3 & 32$\times$4 & 32$\times$1 & 32$\times$4 & 16$\times$1 \\
Number of adapted tokens &3 & 2 & 3 &  3 & 2 & 3 & 3 & 3 & 2\\ 
Training epochs  & 150 & 100& 150 & 150 &100  & 80 & 50 &  100 & 100\\
Best dev performance & 76.2 & 65.4& 90.8 & 89.7 & 93.9 & 92.2 & 83.7 & 87.9 & 94.1\\
Best epochs  & 121 & 63& 109 & 136 & 99 & 42 & 49 & 97 & 89\\
\bottomrule
\end{tabular}
\caption{\modelname with BERT-large training details for GLUE and CoNLL2003 tasks. Distributed training on multiple GPUs is used when avaiable for less training time. }
\label{tab:details}
\end{table*}
Our model is implemented based on Huggingface's Transformers.
We optimize models with AdamW \cite{AdamW}.
We set the maximum input length to 128 and use a fixed random seed of 42 for all tasks.
Experiments are done on NVIDIA RTX A5000 for averagely 3 hours per task, and distributed training is used for most tasks.
\Cref{tab:details} reports the best hyperparameters for model training. For hyperparameter search, we select learning rate from \{5e-4, 1e-3, 2e-3, 2.5e-3, 3e-3, 4.5e-3, 5e-3, 7e-3\} and the number of epochs from \{50, 80, 100, 150\}.

\section{Full Attention Map}
\label{sec:full_map}
\begin{figure*}[t]
    \centering
    \includegraphics[width=0.6\textwidth]{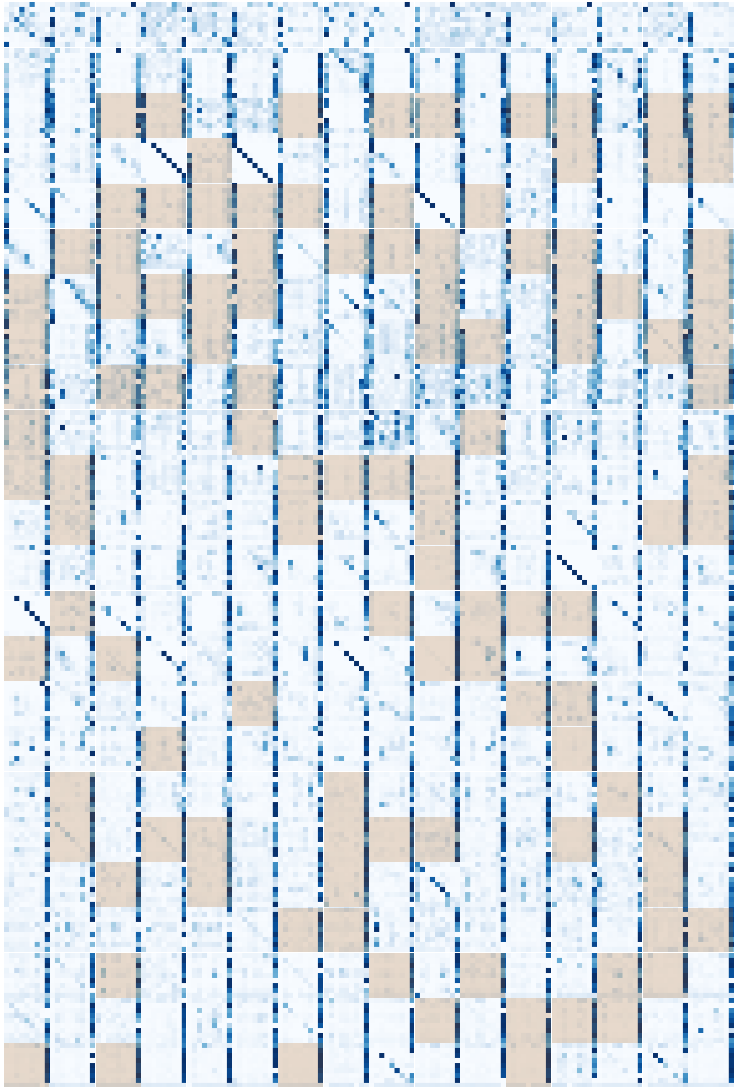}
    \caption{Full attention map of BERT-large pretrained with a random sample from CoLA as input. Rows and columns represent model layers and heads respectively, and darker color indicates larger weights. Vertical attention heads are highlighted in orange.} 
    \label{fig:full_attention_map}
\end{figure*}
\Cref{fig:full_attention_map} illustrates the attention maps of all heads in BERT-large. With a random sample from CoLA dataset as input (\textit{"Fred watered the plants flat."}), there are 112 heads out of 384 in total being vertical heads\textsuperscript{\ref{fn:vertical}}.

\end{document}